\documentclass[letterpaper,10pt,conference]{ieeeconf}
\IEEEoverridecommandlockouts

\usepackage{cite}
\usepackage{amsmath,amssymb,amsfonts}
\usepackage{algorithmic}
\usepackage{graphicx}
\usepackage{textcomp}
\usepackage{url}
\def\BibTeX{{\rm B\kern-.05em{\sc i\kern-.025em b}\kern-.08em
    T\kern-.1667em\lower.7ex\hbox{E}\kern-.125emX}}
\usepackage{nolbreaks}
\usepackage{import}
\usepackage[caption=false, font=footnotesize]{subfig}

\newcommand{\m}[1]{\begin{bmatrix}#1\end{bmatrix}}
\mathchardef\mhyphen="2D

\begin{document}
\title{Feasible Wrench Set Computation for Legged Robots

\thanks{The authors are with the Department of Biomechanical Engineering, University of Twente, Enschede, The Netherlands. This work is part of the research program Wearable
Robotics, project number P16-05, (partly) financed by the
Dutch Research Council (NWO).}
\thanks{© 2022 IEEE. Personal use of this material is permitted. Permission from IEEE must be obtained for all other uses, in any current or future media, including reprinting/republishing this material for advertising or promotional purposes, creating new collective works, for resale or redistribution to servers or lists, or reuse of any copyrighted component of this work in other works.
}
}

\author{{Ander Vallinas Prieto, Arvid Q.L. Keemink, Edwin H.F. van Asseldonk and Herman van der Kooij}

}

\maketitle

\begin{abstract}
 During locomotion, legged robots interact with the ground by sequentially establishing and breaking contact. The interaction wrenches that arise from contact are used to steer the robot's Center of Mass (CoM) and reject perturbations that make the system deviate from the desired trajectory and often make them fall. The feasibility of a given control target (desired CoM wrench or acceleration) is conditioned by the contact point distribution, ground friction, and actuation limits. In this work, we develop a method to compute the set of feasible wrenches that a legged robot can exert on its CoM through contact. The presented method can be used with any amount of non-co-planar contacts and takes into account actuation limits and limitations based on an inelastic contact model with Coulomb friction. This is exemplified with a planar biped model standing with the feet at different heights. Exploiting assumptions from the contact model, we explain how to compute the set of wrenches that are feasible on the CoM when the contacts remain in position as well as the ones that are feasible when some of the contacts are broken. Therefore, this method can be used to assess whether a switch in contact configuration is feasible while achieving a given control task. Furthermore, the method can be used to identify the directions in which the system is not actuated (i.e. a wrench cannot be exerted in those directions).  We show how having a joint be actuated or passive can change the non-actuated wrench directions of a robot at a given pose using a spatial model of a lower-extremity exoskeleton. Therefore, this method is also a useful tool for the design phase of the system. This work presents a useful tool for the control and design of legged systems that extends on the current state of the art.

\end{abstract}

\section{Introduction} 
Preventing dynamic instability during the locomotion of legged robots is of utmost importance. Dynamic instability can cause loss of balance and make the robot fall. Several works have developed stable walking trajectory generators using the concepts of Zero-Moment Point (ZMP)  \cite{VUKOBRATOVIC2004ZERO-MOMENTLIFE}, extrapolated Center of Mass \cite{Hof2005TheStability} and its extensions (Instantaneous Capture Point \cite{Koolen2012Capturability-basedModels} and Divergent Component of Motion \cite{Englsberger2015Three-DimensionalMotion}). The work of Dai et al. \cite{Dai2016PlanningOptimization} used the Contact Wrench Cone (CWC) \cite{Hirukawa2006AZMP}, a generalization of the ZMP criterion to multi-contact on uneven terrain, in order to improve the robustness of walking trajectories in humanoids. They improve the robustness of locomotion by generating trajectories that maximize the magnitude of perturbations that can be rejected. {Caron et al. \cite{Caron2015LeveragingDynamics} develop a more efficient computation of the CWC by considering the wrench generated by each contact surface and making an efficient use of the double description method \cite{Fukuda1996DoubleRevisited}. They use the CWC to find the time-optimal path parametrization of a desired Center of Mass (CoM) trajectory such that stable contact is achieved along the trajectory.}

{The work of Audren and Kheddar \cite{Audren20183-DMulticontact} presents algorithms that compute 3D operational space regions in which the CoM of a system can be located while attaining stable contact for any arbitrary multicontact configuration. Furthermore, the computed stability regions are robust in the sense that a provided set of CoM acceleration perturbations should be rejected without the need to change contact configuration. In \cite{Samadi2021HumanoidContacts}, Samadi et al. approximate the 3D robust stability region of \cite{Audren20183-DMulticontact} by its Chebyshev radius to use it in online feedback control. Through a quadratic progam (QP), they determine a robust CoM location target and contact wrench distribution given some desired interaction forces, which include contact sliding without destabilizing the system. The CoM and wrench distribution task are provided as input to a QP whole body controller (WBC). The WBC may fail to perfectly fulfill these tasks at a certain instant but stability is attained in closed loop. Nevertheless, the stability computations shown in \cite{Samadi2021HumanoidContacts} are `blind' to other motion tasks, so in \cite{Roux2021ControlContacts} the same authors speed up the computations of the 3D stability region \cite{Audren20183-DMulticontact} to provide it as a constraint to the WBC instead.}

{It is worth noting that none of the cited works considers whether the computed trajectory or prescribed task is feasible by the actuators of the system. In \cite{Samadi2021HumanoidContacts} and \cite{Roux2021ControlContacts}, the WBC finds an actuator command that conforms to the limits but the tasks do not take these limits into account explicitly. Concerning optimized trajectories, \cite{Caron2015LeveragingDynamics}  and \cite{Dai2016PlanningOptimization} assume that the actuators are powerful enough to perform the computed trajectories so actuation constraints are not included in their optimizations.}

{We are interested in the feasible wrenches and stability regions for lower body exoskeletons. These devices are supposed to move high payloads while being designed to be lightweight and less material budget is spent on actuators in comparison to humanoid robots. Actuation limits may significantly reduce the feasible wrenches and render some motion infeasible. Furthermore, feasible wrenches are dependent on the current configuration. Therefore, we are interested in the explicit effect of actuation (and lack of it) on the feasible centroidal wrench.} 

Some notable publications that explore how the actuation limits influence the feasible wrench or CoM manipulability are the work by Gu et al. \cite{Gu2015FeasibleRobots} and the work by Orsolino et al. \cite{Orsolino2018ApplicationRobots,Orsolino2020FeasibleRegion}. Both \cite{Gu2015FeasibleRobots} and \cite{Orsolino2018ApplicationRobots} introduce different approaches to compute the set of feasible centroidal wrenches/accelerations that comply with contact {and actuation constraints and \cite{Orsolino2020FeasibleRegion} computes an actuation-aware 2D stability region for the CoM in multi-contact scenarios.}

Gu et al. introduce the Feasible CoM Dynamic Manipulability\cite{Gu2015FeasibleRobots}, which is an ellipsoidal approximation of the set of CoM accelerations that a system can achieve through contact. Their work focuses only on the translation components of the CoM dynamics and takes into account static friction, actuation, and ZMP limits for a single foot on flat ground. The latter restricts the use of their analysis to such scenarios.

Orsolino et al. introduce in \cite{Orsolino2018ApplicationRobots} the Actuation Wrench Polytope\footnote{A polytope is an n-dimensional geometric object with flat faces. In this work, we only consider bounded convex polytopes, used to represent a bounded convex subset of n-dimensional space.} (AWP), whose intersection with the CWC yields a new polytope named the Feasible Wrench Polytope (FWP). This polytope is claimed to contain all wrenches that a legged robot can exert on its CoM through contact, and thus, complies with actuation limits and a given contact model. This approach is more complete than the one presented in \cite{Gu2015FeasibleRobots}, because, by using the CWC, the whole body angular dynamics described at the CoM are also taken into account and the algorithm can be applied to any kind of terrain or multi-contact configuration. 

Nevertheless, to reduce computation time, the method from \cite{Orsolino2018ApplicationRobots} decouples the dynamics of the floating base\footnote{Robots that do not have a base link that is fixed, are free to move through space. This freedom of motion is regarded as a 6 degree of freedom `joint' (3 rotations and 3 translations)} and each branch of the legged system. Furthermore, it takes the actuator torques and generalized accelerations of the system as inputs to compute the contact forces at each leg. However, ground reaction forces are \textit{reactive}, so the contact forces and generalized acceleration are fully determined by the system state and actuation input. As in \cite{Orsolino2018ApplicationRobots} dynamics are decoupled, {their algorithm yields different contact forces for the same actuation input with different input accelerations. Furthermore, the found forces may neither conform to floating base dynamics, nor cause the input accelerations because this part of the dynamics is not enforced. Therefore, this decoupling introduces physical inconsistencies in the resulting polytope.}

{In \cite{Orsolino2020FeasibleRegion}, Orsolino et al. compute a horizontal 2D region in which the projection of the CoM can be stably located under \textit{static} conditions. For these computations, they use the same decoupling of the dynamics as in \cite{Orsolino2018ApplicationRobots} to account for actuation limits but they also constrain floating base dynamics, overcoming the aforementioned inconsistency. This region is valid in the neighborhood of the system configuration for which it was computed and assumes static poses. By recomputing this region, not only do they manage to have a quadruped walk on challenging terrain, but also update the desired trajectories online to prevent loss of balance. The achieved locomotion is, however, quasi-static and relies on a down-scaled version of the computed stability region to guarantee robustness against possible assumption-induced errors. The static assumption is a complete constraint on the joint accelerations and the method we propose constrains accelerations only at contact point level, which allows to compute for wrenches that elicit joint accelerations as well.}

In this work, we present a method to compute the set of feasible wrenches that a robot can exert on its CoM through actuation, which is dependent on pose and contact scenario. The presented method does not decouple system dynamics nor require generalized accelerations as an input, which solves the shortcomings from \cite{Orsolino2018ApplicationRobots} and the resulting set is a more accurate FWP. Our method uses the knowledge of the contact model to constrain the generalized accelerations in such a way that, as dependent variables, they can be substituted in the system of equations and {all system constraints can be expressed in terms of contact forces and actuation torques.} Furthermore, changes in contact configuration are discrete events present in locomotion that further constrain feasible wrenches and we explore the effect of breaking contact on the feasible polytope. 

The main contribution of this paper is the use of the contact model knowledge to build a method that computes the set of feasible wrenches of a system at any given pose. The polytope obtained is the FWP of a robot that maintains the contact configuration. The second contribution of the paper is the extension of the method to compute the set of feasible wrenches that results from opening some contacts. Push-off motions or complete foot detachment are two examples in which this extension is relevant. We apply the method to analyze multi-contact scenarios, contact opening and the presence of passive joints in a system. {Furthermore, we compare our results with the output of \cite{Orsolino2018ApplicationRobots} for a point-feet biped in the plane to show the outcome of not enforcing floating base dynamics in the FWP computation.}

The paper is structured as follows: Sec. \ref{Sec:Method} presents  the proposed method to compute a physically {more accurate} FWP. Sec. \ref{Sec:Examples} shows two examples in which the method is applied: in Sec. \ref{Sec:Planar} the method is applied to a planar model of a biped in uneven contact configuration and in Sec. \ref{Sec:Spatial} the method is used to compute the  FWP of a spatial model of an under-actuated lower-body exoskeleton. Sec. \ref{Sec:Discussion} discusses the attained results and Sec. \ref{Sec:Conclusion} concludes the work.

\section{Method}\label{Sec:Method}
In this section, we present a method to compute the {set of wrenches that a system can exert on its CoM, without violating friction constraints and actuator limits, given the system state and knowledge of contact locations.} 

\subsection{Legged system model}

The Equations of Motion (EoM) of an articulated system are:
\begin{equation}
    M(q)\dot{v}+h(q,v)=S_au+\tau_{e}, \label{Eq:EoM}
\end{equation}
where $q$ is the vector of configuration variables and $v$ is the vector of velocity variables\footnote{The size of $q$ is bigger than the size of $v$ if a quaternion or $SO(3)$ element is used to represent the orientation of the floating base. In such a case $\dot{q}\neq v$.}. Furthermore, $M(q)$ is the configuration-dependent mass-matrix and $h(q,v)$ is the vector of generalized forces due to gravity and velocity effects. The vector $u$ consists of the actuation input (joint torques) and the selector matrix $S_a$ maps the input into generalized forces. External generalized forces acting on the system are represented by $\tau_{e}$. The only external forces considered in this work are contact reaction forces. 

External forces can be modeled as forces $f_i$ acting on a body-fixed point of the system $p_i$. Contact forces arise when any $p_i$ is in contact with the environment. For simplicity, it is common to pre-define contact locations (end-effectors, feet edges, fingertips,...) and assess whether the contact is active. We refer to contacts as active when the distance from the predefined contact point and the environment is zero. Active contact forces contribute to the generalized forces in the following manner:  
\begin{equation}
	\tau_e=\!J^T(q)f\! = \!\m{J_1^T(q)\!\!&\! J_2^T(q) \!\!& \cdots &\!\! J_{n_{k}}^T(q)}\!\m{{}^{1}\!f_1\\{}^{2}\!f_2\\ \vdots \\ {}^{n_{k}}\!f_{n_{k}}},
\end{equation}
where $n_{k}$ is the number of active contacts and $J_i(q)$ is the Jacobian of contact point $p_i$ in frame\footnote{In this work, the coordinate frame $*$ is written as $\psi_*$. If a point, vector or screw is represented in $\psi_*$, this is denoted by the left superscript ${}^*\square$.} $\psi_{i}$:
\begin{equation*}
    J_i(q)=\frac{\partial\left({}^{i}p_i(q)\right)}{\partial q}.
\end{equation*}
For brevity, we will drop the dependency of the aforementioned matrices and vectors on $q$ and $v$.

The coordinate frames used in this work are the inertial frame $\psi_0$, centroidal frame $\psi_c$, which shares orientation with $\psi_0$ but has the origin at the CoM of the system ($x_{c}$), and contact frames $\psi_{i}$, with the origin at a point on the environment closest to $p_i$ (or the foot as a whole{, when more than one contact point is defined in the foot edges}) and with the axes aligned with the tangential and normal directions of the environment{.} 

Looking at \eqref{Eq:EoM} we see that given some state $q,v$ and actuation input $u$ we can solve for generalized accelerations $\dot{v}$ once the external forces are known. However, contact forces are \textit{reactive} and cannot be determined just from \eqref{Eq:EoM}. Thus, 
\begin{equation*}
    \m{M & -J^T}\m{\dot{v}\\f}=S_{a}u-h
\end{equation*}
is underdetermined. Therefore, a contact model is needed.

\subsection{Contact model}

In this work contact is modeled as purely inelastic,as in \cite{Hwangbo2018Per-ContactDynamics}. Thus, contacts follow Signorini's conditions: 
\begin{equation}
    {}^{k}p_N, {}^{k}\!f_N\geq0,\,{}^{k}\!f_N\:\!{}^{k}p_N=0,
\end{equation}
where $\square_N$ represents the normal direction in the contact frame. Notice that we have dropped the contact point index $i$ and we use {a general} contact frame $\psi_k$ because the contact model applies to all contact points.

{For an active contact (${}^{k}p_N = 0$), }{w}e can rewrite Signorini's conditions to velocity and acceleration constraints: 
\begin{equation}
    {}^{k}\dot{p}_N, {}^{k}\ddot{p}_N,{}^{k}\!f_N\geq0,\:{}^{k}\!f_N\:\!{}^{k}\dot{p}_N=0,{}^{k}\!f_N\:\!{}^{k}\ddot{p}_N=0.
\end{equation}

We will also consider Coulomb's friction model, which constrains tangential forces:
\begin{equation}
    \|{}^{k}\!f_T\|_2\leq \mu\:{}^{k}\!f_N, \label{Eq:Friction}
\end{equation}
where $\mu$ is the static friction coefficient, and $\square_T$ is used to denote the tangential direction in the contact frame. 

To completely constrain contact forces we adhere to the maximum dissipation principle: contact forces minimize {the apparent kinetic energy of} active contacts \cite{Hwangbo2018Per-ContactDynamics}.

\subsection{Feasible Wrench Set computation}
For each active contact point {we consider }three possible scenarios\cite{Hwangbo2018Per-ContactDynamics}: 
\begin{enumerate}
    \item \textit{Opening Contact}: When ${}^{k}\dot{p}_N>0$ (or ${}^k\dot{p}_N=0$ and ${}^k\ddot{p}_N>0$), the contact is \textit{opening}, in which case the contact becomes inactive and there is no reaction force:  $\|{}^k\!f\|_2=0$.
    \item \textit{Stick Contact}: {If the contact is not opening, as t}he contact \textit{cannot} penetrate the environment we have that: ${}^k\dot{p}_N=0$, ${}^k\ddot{p}_N=0$. According to the maximum dissipation principle {the apparent kinetic energy at ${}^kp$} must be minimal and, as long as \eqref{Eq:Friction} holds, the global minimum is attained at $\|{}^k\dot{p}\|_2= 0$. If the contact is completely static to start with, the contact forces must ensure $\|{}^k\ddot{p}\|_2=0$ and maintain the contact point in place. These forces that maintain the contact point in place will be called \textit{stick} solutions.
    \item \textit{Slipping contact}: When the contact is not opening and the stick solution violates the friction cone constraint, the contact will \textit{slip} ($\|{}^k\dot{p}_T\|_2\neq 0$). 
\end{enumerate}  

If we focus on controlled legged locomotion, it is desirable to know the wrenches that can be applied on the CoM by: 
\begin{itemize}
    \item not changing the contact configuration. These wrenches are the {stick} solutions.
    \item taking a step or producing a push-off motion, which consists in {opening} some active contacts while the rest {stick}. These wrenches are the opening solutions.
\end{itemize} 

Pure {opening} solutions (all active contacts {open}) correspond to jumping, in which no contact force can be applied. {Contact points slip when the force that would make them stick violates the friction constraints. Thus, slip forces belong to friction cone facets and as such, some of these forces will be present in the corresponding facets of the FWP. Explicit knowledge of slip forces} could potentially be used for CoM control but are not considered in this work.

{Our method consists on the following steps:}
\begin{enumerate}
    \item {Construct the half-plane ($\mathcal{H}-$)description \cite{Fukuda1996DoubleRevisited} of $f$ and $u$ (\eqref{Eq:StickAFP}, \eqref{Eq:StickIneq} or \eqref{Eq:OpenAFP}, \eqref{Eq:OpenIneq})}
    \item {Compute the vertex ($\mathcal{V}-$)description of this set using a vertex enumeration algorithm (\eqref{Eq:SVenum} or \eqref{Eq:OVenum})}
    \item {Transform all $f$ vertices into centroidal wrench (${}^cw$) vertices (\eqref{Eq:TFS} or \eqref{Eq:TFO}, in the same fashion as \cite{Orsolino2018ApplicationRobots,Caron2015LeveragingDynamics})}
\end{enumerate}

In the remainder of the section we first introduce our notation for the $\mathcal{H}$ and $\mathcal{V}-$description of a polytope. Then, we briefly refresh how to express all friction cones and actuation limits as linear inequalities and finally derive the constraints that enforce stick forces and opening+stick solutions.
\subsubsection{Vertex and Half-plane Description of Polytopes}
Let $P\subseteq\mathbb{R}^n$ be a bounded convex polytope. $P$ can be described in terms of its vertices and its bounding half-planes: 
\begin{equation}
    \begin{split}
    P&=\left\{x\in \mathbb{R}^{n}:x=V\alpha,\: \underset{l=1}{\overset{m}{\sum}}\alpha_l=1,\:0\leq\alpha_l\leq1\right\},\\
        P&=\{x\in \mathbb{R}^{n}:Ax\leq b\cap Cx=d\}.
    \end{split}
\end{equation}
where $m$ is the number of columns of $V$, which are the vertices of $P$. The rows of $A,b$ and $C,d$ describe, respectively, the inequality and equality constraints that bound $P$. Our notation for the $\mathcal{V}-$ and $\mathcal{H}-$description of $P$ is:
\begin{equation}
    \begin{split}
        \mathcal{V}(P)&=V,\\
        \mathcal{H}(P)&=\left\{A,b,C,d\right\}.
    \end{split}
\end{equation}
Vertex and facet enumeration algorithms compute one representation given the other:
\begin{equation}
    \begin{split}
        \mathcal{V}(P)&:=\text{vertexEnum}(A,b,C,d),\\
        \mathcal{H}(P)&:=\text{facetEnum}(V).
    \end{split}
\end{equation}
In order to compute the set of feasible wrenches, we first construct $\mathcal{H}(P_s)$ and $\mathcal{H}(P_o)$, which are the sets of constraints that bound $x=\m{f^T&u^T}^T$ when all contact points stick and when some points stick and other open\footnote{$x\in\mathbb{R}^{tn_k+n_a}$ with $t=2$ in 2D and $t=3$ in 3D}. Then we compute\footnote{In vertexEnum() we project down the inequalities to the equality-constrained space of $C,d$, compute the vertices in this reduced space (fewer variables) and project them back to $x$.} $\mathcal{V}(P_s)$ and $\mathcal{V}(P_o)$, and finally we transform these vertices to centroidal wrench vertices for stick and stick+open solutions respectively.
\subsubsection{Actuation limits and Linearized friction cone}
Regardless of contact configuration, we have that $u$ is bounded on each entry $r$:
\begin{equation}
    u_r \in [u_r^-,u_r^+],
\end{equation}
with $u_r^-<0<u_r^+$. Thus, the actuation space can be regarded as a convex polytope whose $\mathcal{H}-$description is given by:
\begin{equation}
	\m{\mathbb{I}_{n_{a}}\\-\mathbb{I}_{n_{a}}}\m{u_1\\ \vdots \\ u_{n_{a}}}=A_{a}u\leq \m{u_1^+\\ \vdots\\u_{n_{a}}^+\\-u_1^-\\ \vdots\\ -u_{n_{a}}^-}= b_{a},
	\label{Eq:TLim}
\end{equation}
where $\mathbb{I}$ is the identity matrix (of size equal to the subscript) and $n_a$ equals the number of actuators.

Furthermore, for any given time instant with known active contacts, the friction cone constraint for every ${}^k\!f$ are given by ${{}^k\!f_{N}\geq0}$ and \eqref{Eq:Friction}. In the plane, \eqref{Eq:Friction} is linear and the $\mathcal{H}-$description is given by:
\begin{equation}
    \m{-1 & -\mu\\1 &-\mu\\0 &-1}\m{{}^k\!f_{T}\\{}^k\!f_{N}}=A_{f}{}^k\!f\leq b_{f}=\m{0\\0\\0}.
    \label{Eq:FinP}
\end{equation}
 In 3D, \eqref{Eq:Friction} describes a cone. We have chosen to linearly approximate this cone with the inner pyramidal approximation:
\begin{equation}
     \m{-1 & 0 &  -\frac{\mu}{\sqrt{2}}\\1&0 &-\frac{\mu}{\sqrt{2}}\\0 &-1 &   -\frac{\mu}{\sqrt{2}}\\0 &1&-\frac{\mu}{\sqrt{2}}\\0&0 &-1}\m{{}^k\!f_{T1}\\{}^k\!f_{T2}\\{}^k\!f_{N}}=A_{f}{}^k\!f\leq b_{f}=\m{0\\0\\0\\0\\0}.
     \label{Eq:FinS}
\end{equation}

\subsubsection{Stick constraints}
To find the \textit{stick} solutions we  constrain the acceleration of all active contacts to be zero. Contact point acceleration is expressed in terms of the generalized velocities, accelerations, and the contact Jacobian:

\begin{equation}
    {}^{i}\ddot{p}_i=J_i\dot{v}+\dot{J}_iv=0.
    \label{Eq:StickCT}
\end{equation}

Combining \eqref{Eq:StickCT} with \eqref{Eq:EoM} we get:
\begin{equation}
    \m{M & -J^T\\ J & 0 }\m{\dot{v}\\f}=\m{S_{a}u-h\\-\dot{J}v}. \label{Eq:fullEM}
\end{equation}
As $\dot{v}$ can be written in terms of $x=\m{f^T&u^T}^T$, we find the following \emph{stick} equality constraint:
\begin{equation}
\begin{split}
    \m{JM^{-1}J^T&JM^{-1}S_a}x &= JM^{-1}h-\dot{J}v,\\
    C_{s}x&=d_{s}.
\end{split}\label{Eq:StickAFP}
\end{equation}
Collecting \eqref{Eq:TLim} and \eqref{Eq:FinP} or \eqref{Eq:FinS} for every active contact point we complete the $\mathcal{H}-$description of $x$:
\begin{equation}
\begin{split}
    \m{\text{diag}(A_{f_1},...,A_{f_{n_k}}) & 0\\0&A_a}x&\leq\m{0\\b_a},\\\vspace{3pt}A_{s}x&\leq b_{s},
\end{split}\label{Eq:StickIneq}
\end{equation}
where diag() represents the block-diagonal stacking of the friction constraints. Therefore, we have:
\begin{equation}
\begin{split}
    \mathcal{H}(P_s)&=\left\{A_{s},b_{s},C_{s},d_{s}\right\},\\
    \mathcal{V}(P_s)&:=\text{vertexEnum}\left(A_{s},b_{s},C_{s},d_{s}\right).
\end{split}\label{Eq:SVenum}
\end{equation}
\noindent Constraint \eqref{Eq:StickCT} is bilateral, which means that contact points can neither penetrate the contact surface nor \textit{open}. However, the friction cone constraints will prevent that the contact `pulls' on the ground and limit the tangential forces. 

Moreover, constraint \eqref{Eq:StickCT} can only be used with established contacts ($\|{}^{i}\dot{p}_i\|=0$). During impact events, the contact points have non-zero velocity, and therefore \eqref{Eq:StickCT} cannot be applied to compute the contact forces. In this paper, we will only address established contacts. 

\subsubsection{Opening constraints}

In order to compute forces that are feasible when some of the active contacts \textit{open}, the reasoning is the following:

\begin{enumerate}
    \item The set of active points that we choose to \textit{open} produce zero force ($\|f_o\|_2=0$), so we only need to solve for the forces of the active contacts that \textit{stick} ($f_s$)
    \item The set of active points that \textit{stick} cannot accelerate ($\|\ddot{p}_s\|_2 = 0$) and the active contacts that open cannot penetrate the environment ($\ddot{p}_{oN}\geq0$)
\end{enumerate}

\noindent As $\|f_o\|_2=0$, we can remove $f_o$ from $x$ and reduce our variables to $x_o=\m{f_s^T&u^T}^T$. Eq. \eqref{Eq:fullEM} is reduced to constrain the motion of the \textit{stick} contacts only, and account for the force only in those points:
\begin{equation}
    \m{M & -J_s^T\\ J_s & 0 }\m{\dot{v}\\f_s}=\m{S_{a}u-h\\-\dot{J}_sv},
\end{equation}
 where $J_s$ is the stack of contact jacobians corresponding to the active contacts that stick. Thus, the equality constraint on $x_o$ is given by:
\begin{equation}
\begin{split}
    \m{J_sM^{-1}J_s^T&J_sM^{-1}S_a}x_o &= J_sM^{-1}h-\dot{J}_sv,\\
    C_{o}x_o&=d_{o}.
\end{split}\label{Eq:OpenAFP}
\end{equation}
The opening contacts cannot penetrate the environment. Therefore, it is insufficient to constrain the motion of the contacts that stick to compute the FWP\textsubscript{o}; the $f_s$ cannot cause the \textit{opening} contacts to penetrate the environment. In our acceleration formulation, the following must hold:
\begin{equation}
    J_{oN}\dot{v}+\dot{J}_{oN}v\geq0, \label{Eq:nonpCons}
\end{equation}
where $J_{oN}$ represents the normal component of the stack of contact jacobians corresponding to the active points that open. This non-penetration inequality can be rewriten in terms of $x_o$:
\begin{equation}
    \begin{split}
    \m{-J_{oN}M^{-1}J_s^T&-J_{oN}M^{-1}S_a}x_o &\leq \dot{J}_{oN}v-J_{oN}M^{-1}h,\\
    A_{np}x_o&\leq b_{np}.
\end{split}
\end{equation}
These inequalities, along with \eqref{Eq:TLim} and the friction cones \eqref{Eq:FinP} or \eqref{Eq:FinS} corresponding to the contact points that \textit{stick}, complete the inequalities bounding $x_o$:
\begin{equation}
\begin{split}
    \m{
    \begin{matrix}
    \text{diag}(A_{f_1},...,A_{f_{n_s}}) & 0\\0&A_a
    \end{matrix}\\A_{np}}x_o&\leq\m{0\\b_a\\b_{np}}\\\vspace{3pt}A_{o}x_o&\leq b_{o}
\end{split}\label{Eq:OpenIneq}
\end{equation}
and, therefore:
\begin{equation}
\begin{split}
    \mathcal{H}(P_o)&=\left\{A_{o},b_{o},C_{o},d_{o}\right\}\\
    \mathcal{V}(P_o)&:=\text{vertexEnum}\left(A_{o},b_{o},C_{o},d_{o}\right).
\end{split}\label{Eq:OVenum}
\end{equation}
As this constraint only applies in the time frame in which the contact point switches from active to open, it may be tempting to approximate the FWP\textsubscript{o} by the FWP\textsubscript{s} with fewer contact points, which means neglecting \eqref{Eq:nonpCons}. In Section \ref{Sec:Planar} we show the importance of including \eqref{Eq:nonpCons}: disregarding it (naïve computation, n-FWP\textsubscript{o}) yields the wrong FWP\textsubscript{o}, because it overestimates the capabilities of the system.

\subsubsection{Transformation to feasible wrenches}
Any force acting at a contact point can be transformed into the corresponding contact wrench in $\psi_c$ as follows:
\begin{equation}
    {}^cw_i=\m{{}^0\Tilde{p}_i-{}^0\Tilde{x}_{c}\\ \mathbb{I}_3}\!{}^0\!R_i\:{}^i\!f_i=T_i{}^i\!f_i
\end{equation}
where $\Tilde{\square}$ is the matrix-form of the vector cross-product and ${}^0R_i$ is the rotation matrix from $\psi_i$ to $\psi_0$. The total centroidal wrench ${}^cw$ equals the sum of each individual contact wrench:
\begin{equation}
    {}^cw=\m{T_1&\cdots&T_{n_k}}f=Tf
\end{equation}
Thus, we can compute the vertices of the set of feasible wrenches when all contact points stick (FWP\textsubscript{s}):
\begin{equation}
    \mathcal{V}(\text{FWP\textsubscript{s}})=\m{T&0}\mathcal{V}(P_s).\label{Eq:TFS}
\end{equation}
When some contacts open, the force at those points is null so, to compute the feasible wrenches when some of the active contacts open (FWP\textsubscript{o}), we truncate $T$ to only contain the transforms of the points that \textit{stick} ($T_s$):
\begin{equation}
    \mathcal{V}(\text{FWP\textsubscript{o}})=\m{T_s&0}\mathcal{V}(P_o).\label{Eq:TFO}
\end{equation}

For a given contact configuration with $n_k$ contact points there is one FWP\textsubscript{s} and $2^{n_k}-2$ FWP\textsubscript{o}. The union of FWP\textsubscript{s} and all non-empty FWP\textsubscript{o} yields the complete {FWP}, which is the (non-convex) set of wrenches that the system can exert on its CoM through actuation.

\section{Examples} \label{Sec:Examples}
\subsection{FWP of a planar biped model} \label{Sec:Planar}
\begin{figure}
    \vspace{5pt}
    \includegraphics[width=\columnwidth]{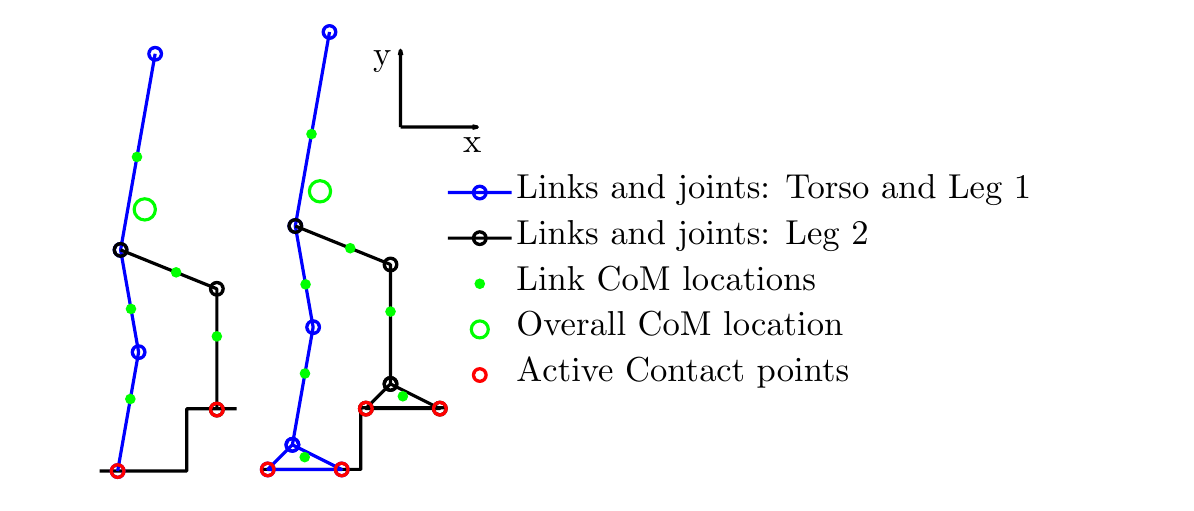}
    \vspace{-25pt}
    \caption{Biped planar models with both feet in contact on uneven terrain. Left: point-feet model Right: triangular-feet model}
    \vspace{-15pt}
    \label{fig:Stick_Stairs}
\end{figure}

To illustrate the added value of our method we apply it to a planar model of an anthropomorphic biped climbing stairs. When a biped climbs stairs (as shown in Fig. \ref{fig:Stick_Stairs}), it confronts multi-contact situations (double support phase) on uneven terrain (contacts are not 
at the same height), in which case co-planar contact simplifications cannot be used. In Fig. \ref{fig:Stick_Stairs} we show two bipeds, the first with point-feet and the second with triangular-feet and ankle joints. The point-feet biped model consists of a torso and two legs with hip and knee, and the segment length and inertial properties are computed for the standardized model by Winter \cite{Winter2009BiomechanicsEdition} for a human that is 1.7 m tall and weighs 70 kg. The triangular-feet model is identical to the point-feet model when the feet ($m_f=0.8$ kg, $h_f=0.1$ m, $l_f=0.3$ m) and ankle joints are added.

We first compute the FWP\textsubscript{s} of the point-feet biped with our method and compare it with the output of the method from \cite{Orsolino2018ApplicationRobots}, which works for point-feet and would require further computations otherwise. We solve for two forces per leg\footnote{We compute their version of the FWP applying equations (3), (10-13) from \cite{Orsolino2018ApplicationRobots}, setting $\dot{v}=0$.}, where each leg has two actuators (hip and knee). All results shown in this work were attained using Matlab 2021b. All polytope intersections and double-description operations in this work were computed using \cite{MPT3}. 

\begin{figure}
    \vspace{10pt}
    \includegraphics{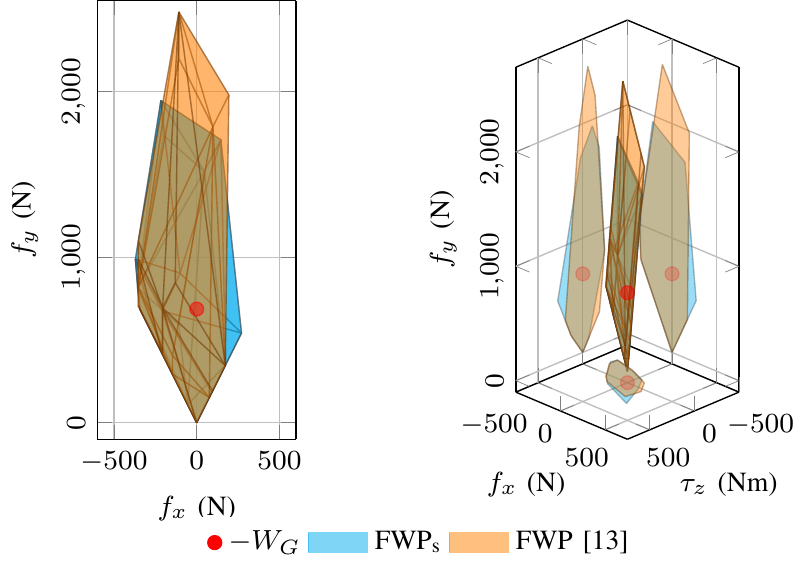}
    \caption{FWP\textsubscript{s} and FWP according to \cite{Orsolino2018ApplicationRobots} of the point-feet biped model in the configuration depicted in Fig. \ref{fig:Stick_Stairs} and gravity compensation wrench ($-W_G$) with their projections in the principal back-planes. Left: the FWP\textsubscript{s} is presented as the projection of the whole polytope in the linear force plane, similar to the results analyzed in \cite{Gu2015FeasibleRobots}. Right: the complete wrench space is shown, and the feasible moments around the CoM are also visible. All wrenches are expressed in $\psi_c$: $x$-axis corresponds to the horizontal direction, $y$-axis to the vertical and the $z$-axis is the direction out of the page (counter-clockwise rotation as positive).
    }
    \label{fig:FWPs_Stairs}
    \vspace{-15pt}
\end{figure}

With a friction coefficient $\mu=0.5$ and torque limits $u^\pm=\pm100\: \text{Nm}$ for all joints, the FWP\textsubscript{s} of the point-feet biped model at the configuration shown in Fig. \ref{fig:Stick_Stairs} and zero joint velocity (computed both with our method and with the algorithm from \cite{Orsolino2018ApplicationRobots}) is presented in Fig. \ref{fig:FWPs_Stairs}. On the left, the linear forces are shown. The friction cone constraint produces a sharp tip at zero force and follows the linear slope defined by $\mu$, which is clearly depicted in this figure. On the right, the full FWP\textsubscript{s} is shown. Fig \ref{fig:FWPs_Stairs} shows with the complete polytopes and their side projections that also a moment may be applied around the CoM. As the polytopes have non-zero volume, the system can exert the same force on the CoM with various resulting moments, because the force distribution can be divided between the feet. Although these observations apply to both our FWP and the one computed using the method from \cite{Orsolino2018ApplicationRobots}, there are differences. The latter method underestimates the maximum horizontal forces that the system could exert at the CoM, while the maximum vertical force is overestimated. Thus, some of the wrenches computed with the method from \cite{Orsolino2018ApplicationRobots} are not feasible.  

\begin{figure}
    \vspace{10pt}
    \includegraphics{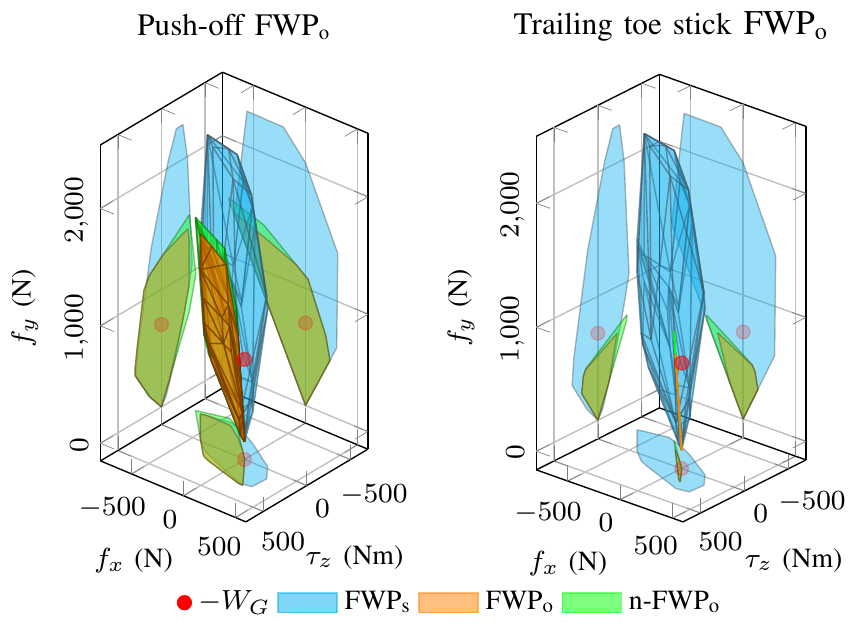}
    \caption{Gravity compensation wrench ($-W_G$), FWP\textsubscript{s} and two different FWP\textsubscript{o} (and n-FWP\textsubscript{o}) of the triangular-feet biped model in the configuration depicted in Fig. \ref{fig:Stick_Stairs} with their projections in the principal back-planes. Left FWP\textsubscript{o}: all points stick except the heel of the trailing leg. Right FWP\textsubscript{o}: the toe of the trailing leg sticks and the other points open. All wrenches are expressed in $\psi_c$. n-FWP\textsubscript{o} corresponds to the naïve computation of the FWP\textsubscript{o}, which does not take \eqref{Eq:nonpCons} into account.
    }
    \label{fig:FWPs_2FWPo_Stairs}
    \vspace{-10pt}
\end{figure}

As presented in Sec.\ref{Sec:Method}, the FWP\textsubscript{s} only contains the wrenches that are feasible when all the contacts stick. Fig. \ref{fig:FWPs_2FWPo_Stairs} shows the FWP\textsubscript{s} superposed with  FWP\textsubscript{o} for two different contact opening situations for the triangular-feet biped model (with $\mu=0.5$ and $u^\pm=\pm100$ for all joints). Furthermore, we show (in green) the naïve-FWP\textsubscript{o}, which disregards \eqref{Eq:nonpCons} in the computations. The first FWP\textsubscript{o}, on the left, is the set of feasible wrenches if a push-off motion is started (the heel of the trailing leg detaches while the toe remains in contact). This polytope shows that a (slightly) greater positive moment can be applied on the CoM when the heel detaches than when the whole foot remains in place, while the feasible force in the positive horizontal direction is reduced. The system is statically unstable in this opening contact scenario because static gravity compensation is not feasible.
However, with the push-off, the biped can compensate gravity while some added moment is applied around the CoM. Thus, the resulting linear acceleration can be zero but in such a case the whole-body angular momentum rate will not be zero. 

The second FWP\textsubscript{o} (Figure \ref{fig:FWPs_2FWPo_Stairs}, right panel)  corresponds to the situation in which all points except the toe of the trailing leg open. This is a more challenging situation and all feasible forces will have an associated moment that cannot be independently modulated. Therefore, any desired wrench outside of this \nolbreaks{(hyper-)plane} (be it desired change in centroidal momentum or gravity/disturbance compensation) is not feasible. Only its projection onto the FWP\textsubscript{o} is feasible and therefore the system will accelerate in the non-compensated direction. 

To conclude, neglecting \eqref{Eq:nonpCons} yields some n-FWP\textsubscript{o} that differ from our proposed FWP\textsubscript{o}. The found  differences are not too big and the n-FWP\textsubscript{o} could be considered an outer approximation of the FWP\textsubscript{o}. Nevertheless, failing to comply with \eqref{Eq:nonpCons} means that the contact opening will not happen. 

\subsection{FWP of a three-dimensional exoskeleton model: Comparison of passive and active DoF} \label{Sec:Spatial}
In this section, we apply the method to a spatial model, where the centroidal wrench has 6 components (3 moments and 3 forces). Due to the higher dimension of this space, the polytopes cannot be easily visualized and the results are more challenging to interpret. Nevertheless, the FWP is a useful tool to evaluate the design of an articulated system. 

To exemplify this we use a rigid multi-body mathematical model of our Symbitron exoskeleton \cite{Meijneke2021SymbitronIndividuals}, which is a lower-body exoskeleton with 6 joints per leg: hip endo-exo rotation, hip abduction-adduction, hip, knee, and ankle flexion-extension, and ankle in-eversion. The hip endo-exo rotation is spring-loaded and the ankle in-eversion is non-actuated. The non-actuated joints can be locked and we restrict this analysis to a model in which the hip endo-exo rotation is locked. Therefore, the exoskeleton model has a 6 DoF floating-base joint and 10 revolute joints, 2 of which are passive (we remove the $u_r$ of passive joints from $u$ and the corresponding columns from $S_a$). We use the same $\mu$ and $u{^\pm}$ (in the actuated joints) as Sec. \ref{Sec:Planar}.

\begin{figure}
\vspace{15pt}
    \centering
    \subfloat[\label{fig:ExoA}]{
        \includegraphics[width=0.46\columnwidth]{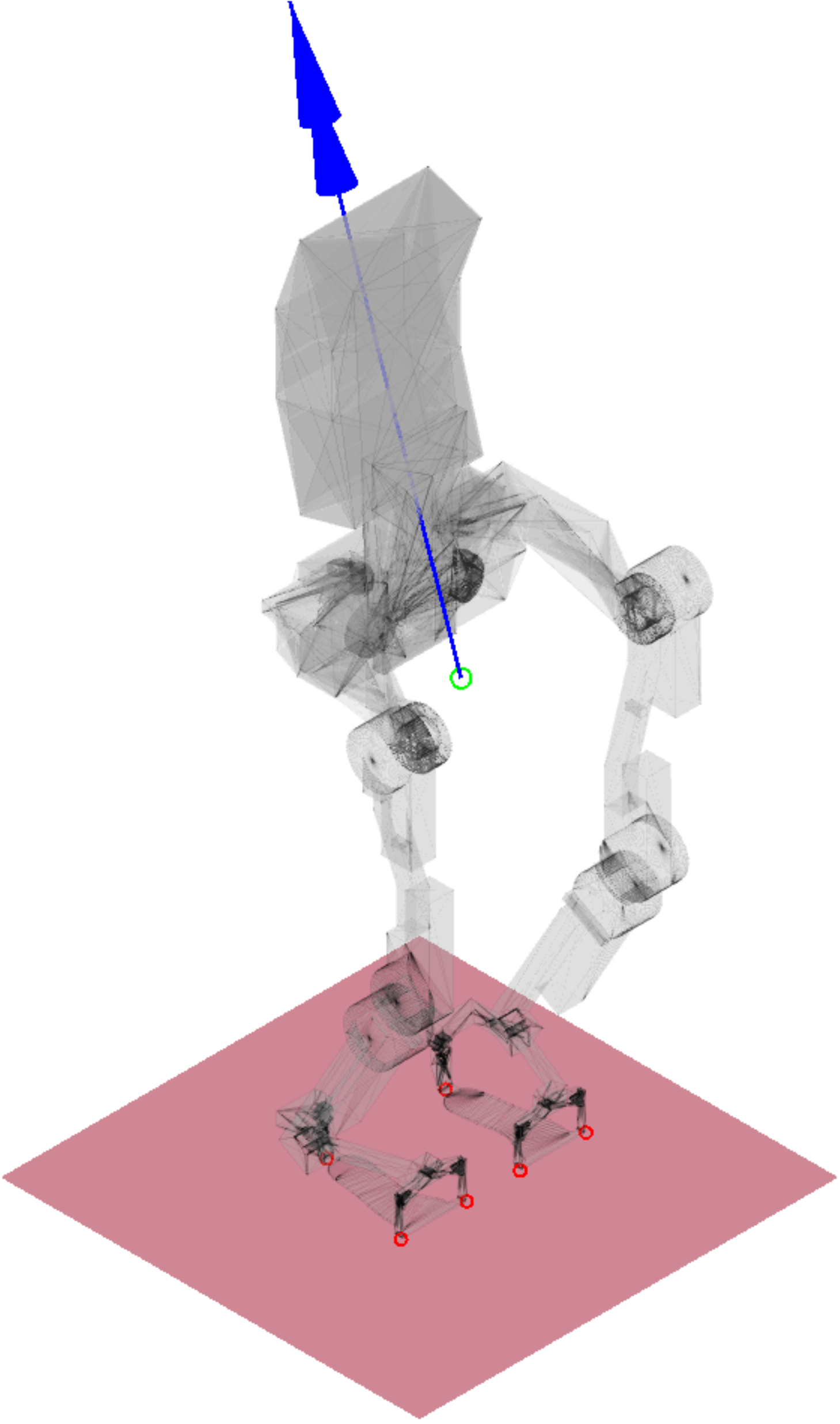}
    }
    \hfill
    \subfloat[\label{fig:ExoB}]{
        \includegraphics[width=0.46\columnwidth]{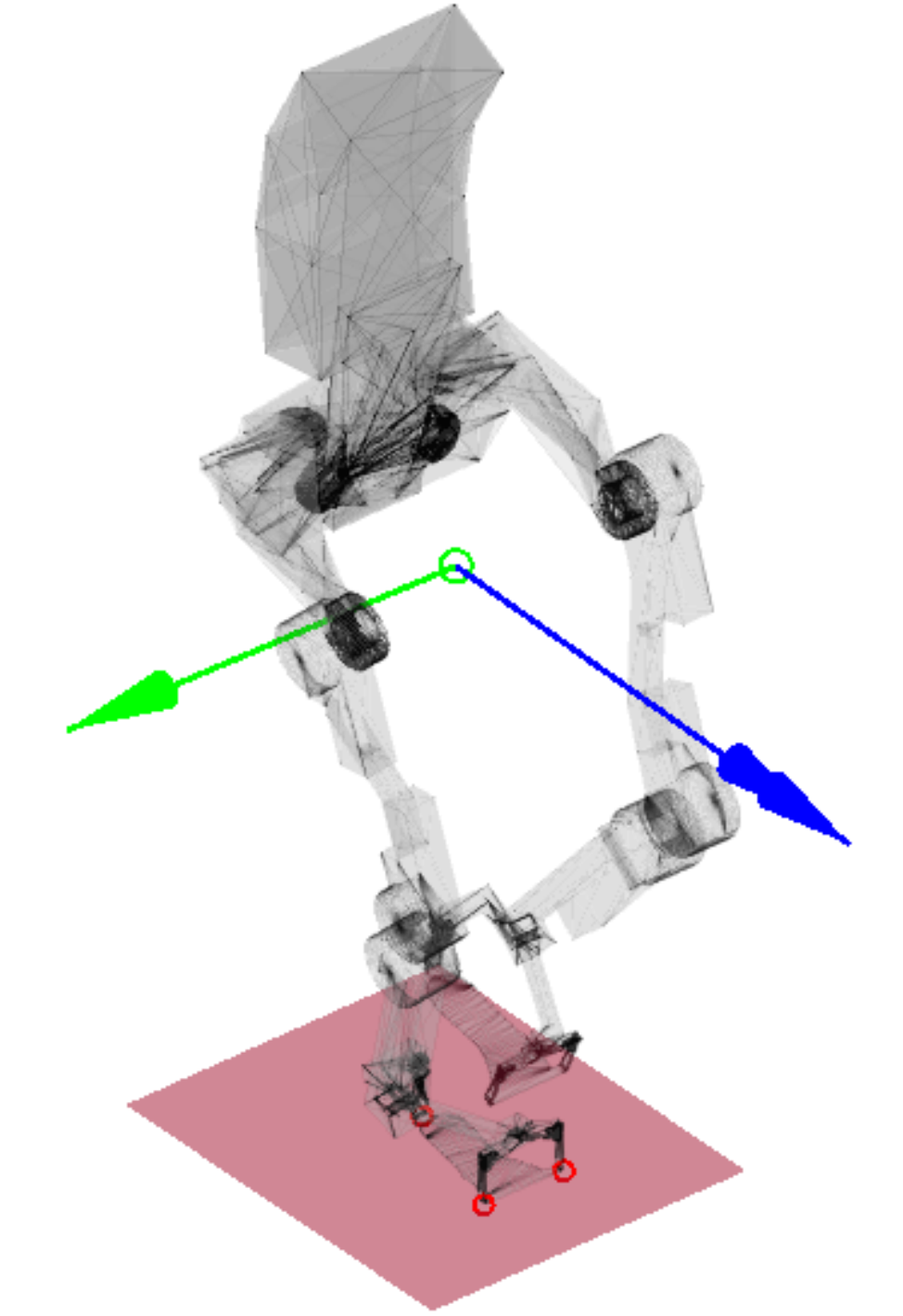}
    }
    \caption{Model of the Symbitron exoskeleton in two different configurations and the corresponding singular wrench directions. The green marker represents the location of the CoM, the red markers show the active contact point locations. (a) Parallel feet double support stance with a little crouch. The blue (double headed) vector represents the non-actuated direction. A moment around this vector is not feasible with the depicted contact configuration. (b) Single support stance in static equilibrium. The blue (double headed) and green (single headed) vectors represent the six components of a unitary CoM wrench in the non-actuated direction. A moment around the blue arrow combined with a force along the green arrow is not feasible (where the moment/force proportion is given by the vector lengths) with the depicted contact configuration if the ankle in-eversion joint is passive.}
    \label{fig:3DDSConfig}
    \vspace{-15pt}
\end{figure}

The FWP\textsubscript{s} of the system in the configuration shown in Fig. \ref{fig:ExoA} determines that the exoskeleton is in static equilibrium configuration as gravity compensation is feasible. However, the FWP\textsubscript{s} is singular as it is contained in a 5D hyper-plane\footnote{The singular dimensions correspond to the rows of $C$ in $\mathcal{H}(\text{FWP\textsubscript{s}})$. However, as we were not interested in the complete $\mathcal{H}(\text{FWP\textsubscript{s}})$, the singular directions shown in this section were found by determining the eigenvalues and eigenvectors of the covariance matrix of $\mathcal{V}(\text{FWP}_{s})$. Zero eigenvalues indicate that there is a dimension of zero thickness. The corresponding eigenvector is the direction of zero thickness, i.e. the non-actuated direction.}, instead of being 6D, similar to the FWP\textsubscript{o} shown in Fig. \ref{fig:FWPs_2FWPo_Stairs} being a 2D plane in 3D wrench space. Thus, there is a wrench direction in which the system is not actuated. The singular (non-actuated) direction for this configuration is a pure moment around the axis shown by the blue vector in Fig. \ref{fig:ExoA}. Any actuation input that maintains the contact configuration will elicit the same moment around this axis so the whole-body angular momentum rate of change in this direction is constant (in the absence of perturbations). As the system is in static equilibrium, the feasible moment around the singular axis is zero. A good example in which this is not the case was shown in Fig.\ref{fig:FWPs_2FWPo_Stairs}, where, regardless of the actuation input, the resulting centroidal wrench will belong to the plane and the moment will be non-zero.

For comparison, we checked the FWP\textsubscript{s} of the exoskeleton at the same configuration, now assuming that the ankle in-eversion was actuated, with the same actuator limits as the rest of the joints. The results show that the system would be non-actuated in the same direction. Therefore, the only way to excite the momenta in this direction (through internal forces) is by changing contact configuration.

The benefit of actuated over passive ankle in-eversion is clearer during single support. The FWP\textsubscript{s} of the exoskeleton at the configuration in Fig. \ref{fig:ExoB} is singular in the direction determined by the moment around the blue arrow and linear force along the green arrow when the ankle in-eversion is passive but it would be full dimensional, which means that all momenta would be independently actuated, if the ankle in-eversion were actuated.

\section{Discussion} \label{Sec:Discussion}
In this work, we developed a method to compute the set of feasible wrenches that a legged robot can exert on its CoM through contact. The proposed method can be used for any contact configuration and takes into account actuation limits and a contact model to determine what wrenches are feasible. 

The presented method is an extended version of centroidal manipulability \cite{Gu2015FeasibleRobots} that includes not only rotational dynamics but can also deal with multi-contact situations in uneven terrain like the one shown in Fig. \ref{fig:Stick_Stairs}. It is worth mentioning, however, that linearized friction cones are an internal approximation of the friction model and the results of our method are conservative in 3D. Nonetheless, the linear approximation can be made arbitrarily good, at the cost of increasing the size of the $\mathcal{H}-$description.

The method also overcomes the physical inconsistencies from \cite{Orsolino2018ApplicationRobots} by completely determining reaction forces and generalized accelerations from the system state and actuation input and, therefore, it yields the a more accurate set of feasible wrenches. 

Furthermore, besides showing the capabilities of the system at a given pose, the presented method can also compute the feasible wrenches when some of the contact points open: the FWP\textsubscript{o}-s. These computations can be used to assess whether a change in contact configuration is feasible while still attaining the desired rate of change in centroidal momentum so they can be useful to trigger contact changes and for trajectory planning. In Fig. \ref{fig:FWPs_2FWPo_Stairs} we see that none of the shown FWP\textsubscript{o} contains the gravity compensation wrench. Thus, the switch to any of these contact configurations will excite centroidal dynamics: any actuation that achieves the opening configuration will apply a moment around the CoM and/or make it accelerate. Although the system is not in a static equilibrium when these contacts open, it can still move, which may align with the desired control task.

The method can also be used for comparison purposes and provide useful information on the actuated vs. non-actuated directions of the centroidal wrench. We have shown how some mechanical choices, like the use of passive joints, can have a big impact on the resulting FWP and this information can be used in the mechanical and mechatronic design of the system. Furthermore, prior knowledge of the non-actuated directions can assist in controller synthesis. Any undesired acceleration in the non-actuated direction may result in a fall and is a good indicator to take a step.

In this paper, for simplicity, the models that have been used as examples have flat rigid feet. Nevertheless, legged robots can have flexible feet, which are compliant and have a continuously varying contact surface. Any foot model (and joint impedance) can be easily included to analyze its effect. The addition of a toe-like joint, for example, is trivial and can be a first approximation of the time-varying contact surface, at the cost of increasing the size of $x$ and the $\mathcal{H}-$description.  

The presented method has, however, one big limitation: the method cannot compute the effect of impacts (end effectors at contact distance with the environment and non-zero speed) on the feasible wrenches. This is particularly relevant during locomotion, where feet come in contact with the ground with non-zero velocity. In regular bipedal gait, the system is unstable in a dynamic sense during the latter part of the leg swing motion (the extrapolated CoM is outside of the BoS\cite{Hof2005TheStability}). Heel strike has a twofold effect on stability: 
\begin{enumerate}
    \item The impact prevents ground penetration so, according to the contact model we consider in this work, it generates a kinematically-ideal instantaneous change in velocity. The magnitude of the impulse that achieves this change can be computed, but, as the event is discrete, the force is infinite.
    \item It increases the BoS and the newly available contact forces render the system dynamically stable again.
\end{enumerate}
This method can only show the difference due to the increase in the BoS but cannot predict the change in foot velocity nor the forces arising from the impact.

In \cite{Orsolino2018ApplicationRobots} the FWP is used to maximize robustness in trajectory optimization with prescribed contact locations and instants and our method can be used for the same purpose. The advantage of using our method to compute the FWP in trajectory optimization is that the robustness metric maximized in \cite{Orsolino2018ApplicationRobots} (a relative measure of the leftover wrench space) would not overestimate the Feasible wrenches. 

The main drawback of our method is that its direct implementation is computationally expensive and therefore, not ready for closed loop control. In literature (\!\!\cite{Orsolino2020FeasibleRegion,Samadi2021HumanoidContacts,Roux2021ControlContacts}), computations are in the order of 1-100 ms at most. Our main focus in this research was not time efficiency and as such, the FWP\textsubscript{s}-s (and FWP \cite{Orsolino2018ApplicationRobots}) in \ref{Sec:Planar} have an average computation time of 10ms, (4  (resp. 8) contact forces and 4  (resp. 6) actuators for the point-feet (resp. triangular-feet) model), but the FWP\textsubscript{s} in Fig. \ref{fig:ExoA} was computed in 3 and 4 minutes when ankles were passive and active respectively (18 contact forces and 8 and 10 actuators respectively). Even if faster software was used for the computations, the number of constraints that our method defines is bigger compared to the mentioned approaches and therefore we expect that it will take longer. Therefore, it would be beneficial to make computations more efficient (e.g. using iterative methods like the ones presented in \cite{Audren20183-DMulticontact,Roux2021ControlContacts}) if our method is used for Trajectory Optimization, like in \cite{Orsolino2018ApplicationRobots}. To use the method in online control the computations need to be more efficient, possibly by approximating the set, as in \cite{Samadi2021HumanoidContacts}, or by training a Neural Network that then can be used for Real-Time control. Nevertheless, the method can be used to analyze a system and aid in controller synthesis.

\section{Conclusion} \label{Sec:Conclusion}
In this work, we present a complete method to compute all the Feasible Wrenches on the CoM of a legged robot given its state, the actuation limits and the assumption of inelastic contact. This method extends on the methods presented in literature and can be used to analyze the strengths and weaknesses of legged robots at different poses. This tool can provide useful information for controller synthesis and assessment of the mechanical design of a system.

\bibliographystyle{IEEEtran}
\bibliography{IEEEabrv,Refs}

\end{document}